\documentclass[12pt]{article}

\usepackage{amsmath,amsthm,amssymb,amsfonts}
\usepackage{graphicx}
\usepackage{subfigure}

\usepackage{url}
\usepackage{multirow}
\usepackage{scicite}
\usepackage{color}

\usepackage{hyperref}

\usepackage{times}

\newenvironment{sciabstract}{%
\begin{quote} \bf}
{\end{quote}}

\newcounter{lastnote}

\title{Judging Chemical Reaction Practicality From Positive Sample only Learning}

\author
{
Shu Jiang,$^{1,2\ast}$ Zhuosheng Zhang,$^{1,2\ast}$ \\
Hai Zhao,$^{1,2\dagger}$ Jiangtong Li,$^{2,3}$ Yang Yang,$^{1,2}$ Bao-Liang Lu,$^{1,2}$ Ning Xia$^{4}$\\
\\
\normalsize{$^{1}$Department of Computer Science and Engineering,
        Shanghai Jiao Tong University,}\\
\normalsize{Shanghai, 200240, China}\\
\normalsize{$^{2}$Key Laboratory of Shanghai Education Commission for Intelligent Interaction, }\\
\normalsize{and Cognitive Engineering, Shanghai Jiao Tong University,}\\
\normalsize{Shanghai, 200240, China}\\
\normalsize{$^{3}$College of Zhiyuan, Shanghai Jiao Tong University,}\\
\normalsize{Shanghai, 200240, China}\\
\normalsize{$^{4}$Chemical.AI, Shanghai, 200240, China}\\
\\
\normalsize{$^\ast$These authors contributed equally to this work.}\\
\normalsize{$^\dagger$Corresponding author. Email:  zhaohai@cs.sjtu.edu.cn}
}

\date{}

\begin{document} 


\baselineskip24pt


\maketitle


\begin{sciabstract}
 Chemical reaction practicality is the core task among all symbol intelligence based chemical information processing, for example, it provides indispensable clue for further automatic synthesis route inference.
 Considering that chemical reactions have been represented in a language form, we propose a new solution to generally judge the practicality of organic reaction without considering complex quantum physical modeling or chemistry knowledge. 
 While tackling the practicality judgment as a machine learning task from positive and negative (chemical reaction) samples, all existing studies have to carefully handle the serious insufficiency issue on the negative samples. We propose an auto-construction method to well solve the extensively existed long-term difficulty. 
 Experimental results show our model can effectively predict the practicality of chemical reactions, which achieves a high accuracy of 99.76\% on real large-scale chemical lab reaction practicality judgment. 
\end{sciabstract}


\section*{INTRODUCTION}

Organic reactions including addition reactions \cite{Casiraghi2011ChemInform}, elimination reaction \cite{Elimination}, substitution reactions \cite{substitution1,substitution2,substitution3}, pericyclic reactions \cite{pericyclic}, rearrangement reactions \cite{rearrangement1,rearrangement2}, redox reaction \cite{redox} have been studied for hundreds of years. 
Owing to the development of organic methodology\cite{Methodology}, hundreds of millions of reactions have been practised and more and more compounds have been produced.
Nevertheless, the mechanism of organic reactions has not been completely understood and the practicality of a new organic reaction still mainly relies on human judge from expertise and the eventual exploratory synthesis verification.

Modeling the organic reactions through physical-level method, such as quantum mechanical modeling, is a traditional way to recognize them, whereas it may lead to over-complicated model with poor informative representation \cite{Computational}, even for simple reaction containing only several atoms, is essentially difficult to model due to the need of considering the combinatorial component arrangement using quantum chemistry method. 
Predicting a complicated chemical reaction under a certain condition is even much more challenging\cite{ShortOfCom}, because it requires considering every transition-state, the combination between molecules and their given environment.

Instead, the latest symbol model for chemical reaction has been proposed, in which chemical elements and molecules are regarded as symbols and reactions are considered as text with chemical information. 
Consequently, most text processing methods including machine learning, especially deep learning, can be applied to the text in chemical language. 
Support Vector Machine (SVM) has been proved to be useful to predict the result of crystallization of templated vanadium selenites \cite{nature}. However, it requires complicated manual feature selection with a basis of necessary chemical knowledge. 
Information retrieval is also an effective way to predict the products of organic reactions \cite{infor1,infor2}, which presents a limited candidate set for ranking.
Continuous representation of molecules \cite{continuous} provides a convenient method to automatically generating chemical structures. 
More recently, some researchers \cite{schwaller2017found} cast the reaction prediction task as a translation problem by introducing a template-free sequence-to-sequence model, trained end-to-end and fully data-driven and achieved an accuracy of 80.1\% without relying on auxiliary knowledge such as reaction rules.
Recently, Abigail Doyle et al. \cite{Ahnemaneaar5169} proposed the random forest algorithm, which can accurately predict the yield of Buchwald Hartwig cross-coupling reactions with many detailed features of materials in reactions, though their computational model can only process a kind of reaction and needs too much information about the reactions.

Existing work using machine learning for chemical information processing falls either relying on strong chemical knowledge source or focusing on specific types of reactions. 
Distinctive from previous studies, we provide a cutting-edge symbol alone model on chemical text of organic reactions from a general background. 
A complete data-driven method is proposed for open type of chemical organic reactions, releasing the inconvenient prerequisite with chemical prior knowledge. 
Without complex parameter setting or manual chemical knowledge based feature selection, our approach can automatically discover the salient features and reaction patterns for effective reaction practicality judgment.

In recent years, natural language processing has popularly adopted embedding representation for text units which is a sort of low-dimensional continuous representation learned from neural networks. 
Following the latest advance of deep language processing, embedding is also used to represent chemical text segments for chemical reaction learning. 
Using a data-driven mode, our model will directly learn from a large scale of available reaction data. 
We use reaction formulation collected from the publications for about 1.7 million reactions. Practicality judgment can be straightforwardly formulized into a discriminative machine learning task over two types of reactions, \emph{positive} and \emph{negative}. However, the latter, negative reactions, are seldom reported in chemical literatures and thus usually hard collected. When quite a lot of positive reactions are collocated with few negative ones, the machine learning models have to struggle on seriously imbalanced training dataset. In this work, we propose an effective chemical rule based method for negative reaction generation to cope such a long-term big challenge. Eventually, given
the reactants and products, our model can accurately judge the reaction practicality. 
Our model pipeline is given as follows. 
We first preprocess the SMILES sequence of each reactant and product in atom-wise and adopt an unsupervised segmentation algorithm to tokenize the resulting text into segments in a natural language processing way. 
Then the text or symbol difference between reactants and products which stands for the reaction steps is extracted and tagged from an edit distance detection operation on both sides of reaction text. 
For an effective representation, all the resulting chemical text segments are presented in an embedding form so that either the reactants or the products can be put into vector representation as well. 
At last, the reactants and the products which are all in vectors are fed to a neural network for practicality learning and judgment.


\section*{METHOD}
\begin{figure}
    \centering
    \includegraphics[width=1\textwidth]{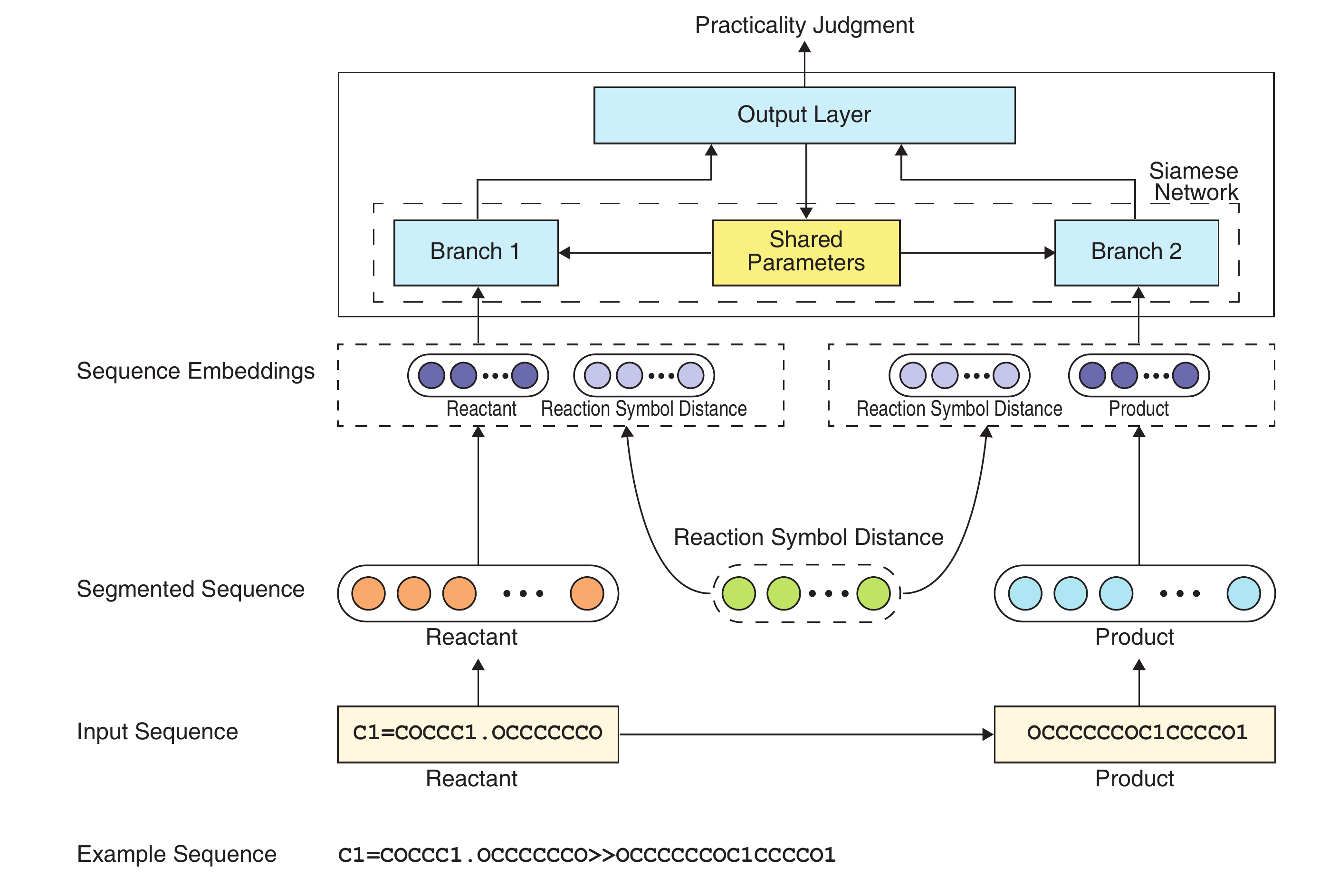}
    \caption{Model for practicality judgment.}
    \label{fig:system}
\end{figure}

For chemical reaction prediction, the key point is to effectively capture the internal relationships between a reactant and the corresponding product representation along with Reaction Symbol Distance (RSD). Note that we assume unreactive reactions will be always kept unreactive under all possible, known reaction conditions, thus we remove all reaction conditions in our judgment.
This task is formulized into a language processing over the corresponding chemical text. 
As shown in Figure \ref{fig:system}, the text segments of reactants and products are represented as vectors of low-dimensional embedding representation. 
Then, a deep neural network is trained to learn the chemical principles of reactions by transforming the feature representation of reactants and products. 
After training, given a reaction text input, the model will judge the practicality.

\subsection*{Unsupervised Tokenization}

SMILES (Simplified Molecular Input Line Entry System) is a line notation for entering and representing molecules and reactions using short ASCII strings, which was initiated by David Weininger at the USEPA Mid-Continent Ecology Division Laboratory in Duluth in the 1980s\cite{Weininger1988SMILES}. 
The primary reason SMILES is more useful than an extended connection table is that it is a linguistic construct, rather than a computer data structure. 
SMILES is a true language, albeit with a simple vocabulary (atom and bond symbols) and only a few grammar rules. 
SMILES representations of structure can in turn be used as \textit{“words”} in the vocabulary of other languages designed for storage of chemical information and chemical intelligence. Some examples are shown in Table \ref{tab:smiles}.

At the very beginning, we remove the hydrogen atoms and the atom mappings from the reaction string, and canonicalized the molecules. 
We treat chemical reaction described by SMILES as a kind of text in natural language. 
Considering that chemical elements (atoms) and various SIMILES bond symbols are \textit{“characters”} in the chemical language, a sequence of SIMILES which stands for chemical compound can be regarded as the corresponding sentence. 
Therefore we need to mine the sequence to find a basic meaningful linguistic unit, word. 
As SMILES encoding text does not provide a word segmentation with solid chemical meaning to facilitate the chemical text processing, we turn to unsupervised tokenization solution in the existing natural language processing \cite{I08-1002}. 
Therefore, we adopt goodness measure based method to tokenize each reactant, product text in SMILES into a sequence of \emph{“words”}. 
Let $W=\{\{w_i,g(w_i)\}_{i=1,...,n}\}$ be a list of character $n$-grams (namely, word candidates) each associated with a goodness score for how likely it is to be a true word from a linguistic/chemical perspective, where $w_i$ is a word candidate and $g(w_i)$ is its goodness function.

The adopted segmentation algorithm is a greedy maximal-matching one with respect to a goodness score.
\begin{equation}\label{eq:equa1}
    \{w^{*},t^{*}\} = \mathop{\arg\max}_{w_1 \ldots w_i \ldots w_n = T} \ \sum_{i=1}^{n} g(w_i)
\end{equation}  
It works on $T$ to output the best current word $w^{*}$ repeatedly with $T=t^{*}$ for the next round as follows, with each $\{w, g(w)\} \in W$.

In our work, we use \emph{Description Length Gain (DLG)} as the goodness measurement for a candidate character $n$-gram from the chemical text. 
In principle, the higher goodness score for a candidate, the more likely it is to be a true word. 
\emph{DLG} was proposed by Kit and Wilks \cite{Kit98thevirtual} for compression-based unsupervised segmentation. 
The DLG extracts all occurrences of $x_{i..j}$ from a corpus $X=x_1x_2...x_n$
and its DLG goodness score is defined as
\begin{equation}\label{eq:dlg}
g_{DLG}(x_{i..j}) = L(X)-L(X[r \to x_{i..j}]  \oplus x_{i..j}),
\end{equation}
where $X[r \to x_{i..j}]$ represents the resultant corpus by replacing all items of $x_{i..j}$ with a new symbol $r$ throughout $X$, and $\oplus$ denotes the concatenation operator.
$L(\cdot)$ is the empirical description length of a corpus in bits that can be estimated by the Shannon-Fano code or Huffman code as below, following classic information theory\cite{Shannon1948},
\begin{equation}\label{eq:shannon}
L(X) \doteq -|X|\sum_{x \in V}\hat{p}(x) \log_{2}\hat{p}(x),
\end{equation}
where $|\cdot|$ denotes the string length, $V$ is the character vocabulary of $X$ and $\hat{p}(x)$ is $x$'s frequency in $X$.

\subsection*{Reaction Symbol Distance (RSD) Generation}
To formally represent the text difference from reactants to products in a reaction formula, we introduce the formal concept of \emph{Reaction Symbol Distance (RSD)}, which indicates how source chemical text can be transformed into target one through a series of symbol inserting and deleting operations. The text operation series can be decoded from calculating the \emph{edit distance} \cite{needleman1970general}.
Edit distance is used to quantify how dissimilar two strings are to one another by counting the minimum number of operations required to transform one string into the other.

For a source sequence $S= s_1s_2 \ldots s_n$ and the target sequence $T= t_1t_2\ldots t_m$, the RSD sequence $R= r_1r_2\ldots r_n$ is encoded by the following tags:
\begin{itemize}
    \item \textbf{AD} indicates a string should be added right before the corresponding location.
    \item \textbf{AR} indicates the corresponding symbol should be replaced by the given string with the tag.
    \item \textbf{RR} the corresponding symbol should be deleted.
    \item \textbf{\_} means that there is no operation at the location.
\end{itemize}

\begin{figure}
    \centering
    \includegraphics[width=1\textwidth]{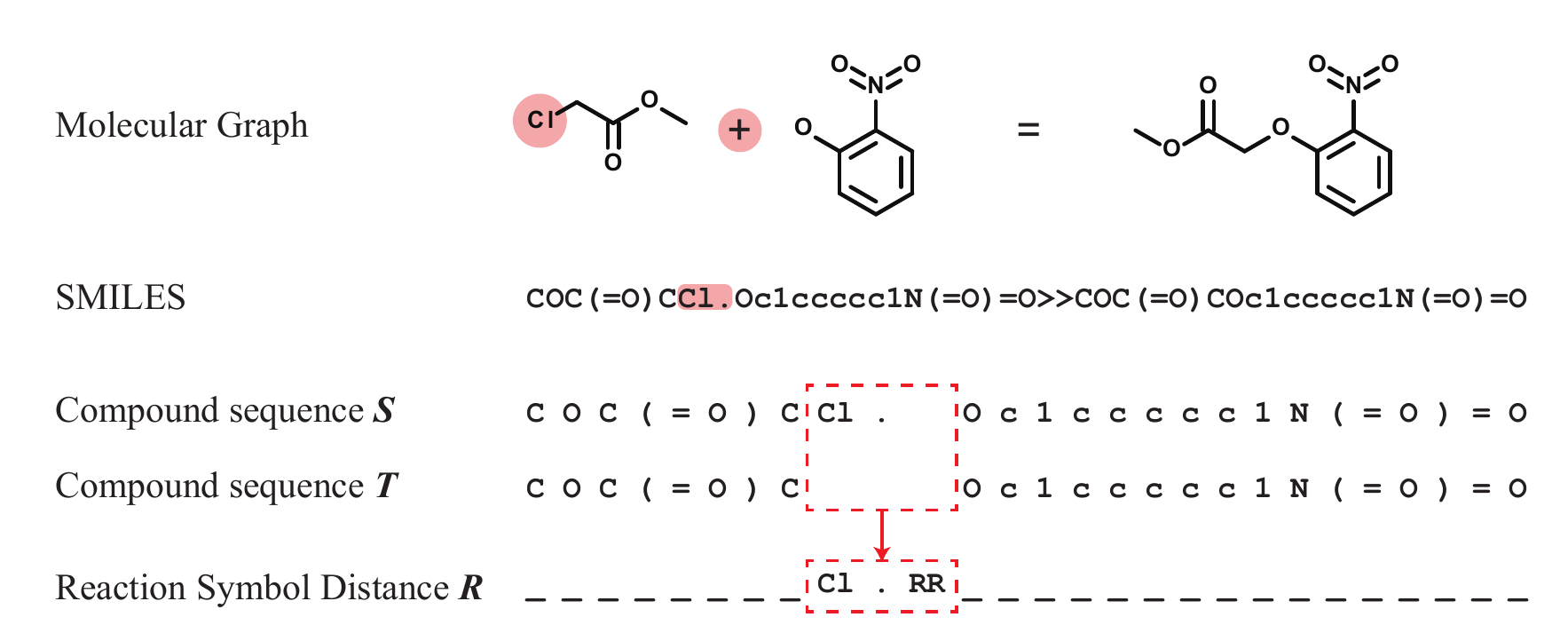}
    \caption{Generation of Reaction Symbol Distance (RSD).}
    \label{fig:smiles}
\end{figure}

All compound sequences $S$ and $T$ are split into elements, and the resulting RSD from $S$ to $T$ are illustrated in the Figure \ref{fig:smiles}.

The data processing steps together with examples are summarized in Table \ref{tab:step}. The same preprocessing steps were applied to all datasets.

\subsection*{Embedding}
In our adopted neural model, an embedding layer is used to map each element or segmented \textit{“word”} from a sequence into a vector with dimension $d$.
Our model takes three types of inputs, reactant, RSD and product. After embedding, the reactant sequence with $n$ words is represented as $\mathbb{R}^{d\times n}$. 
Similarly, we obtain the embeddings of the reactant sequence $\mathbf{R}$, the RSD sequence $\mathbf{S}$ and the product sequence $\mathbf{P}$. 
Then, the input sequences are subsequently aggregated into two compact representations through projection and concatenating:
\begin{align}
\mathbf{M_1}=\begin{bmatrix}
\mathbf{R}_1^1 \oplus \mathbf{S}_1^1 \\
\vdots \\
\mathbf{R}_h^1 \oplus \mathbf{S}_h^1 \\
\end{bmatrix}
,
\mathbf{M_2}=\begin{bmatrix}
\mathbf{P}_1^1 \oplus \mathbf{S}_1^1 \\
\vdots \\
\mathbf{P}_h^1 \oplus \mathbf{S}_h^1 \\
\end{bmatrix}
\end{align}

\subsection*{Siamese Network}
\begin{figure}
    \centering
    \includegraphics[width=1\textwidth]{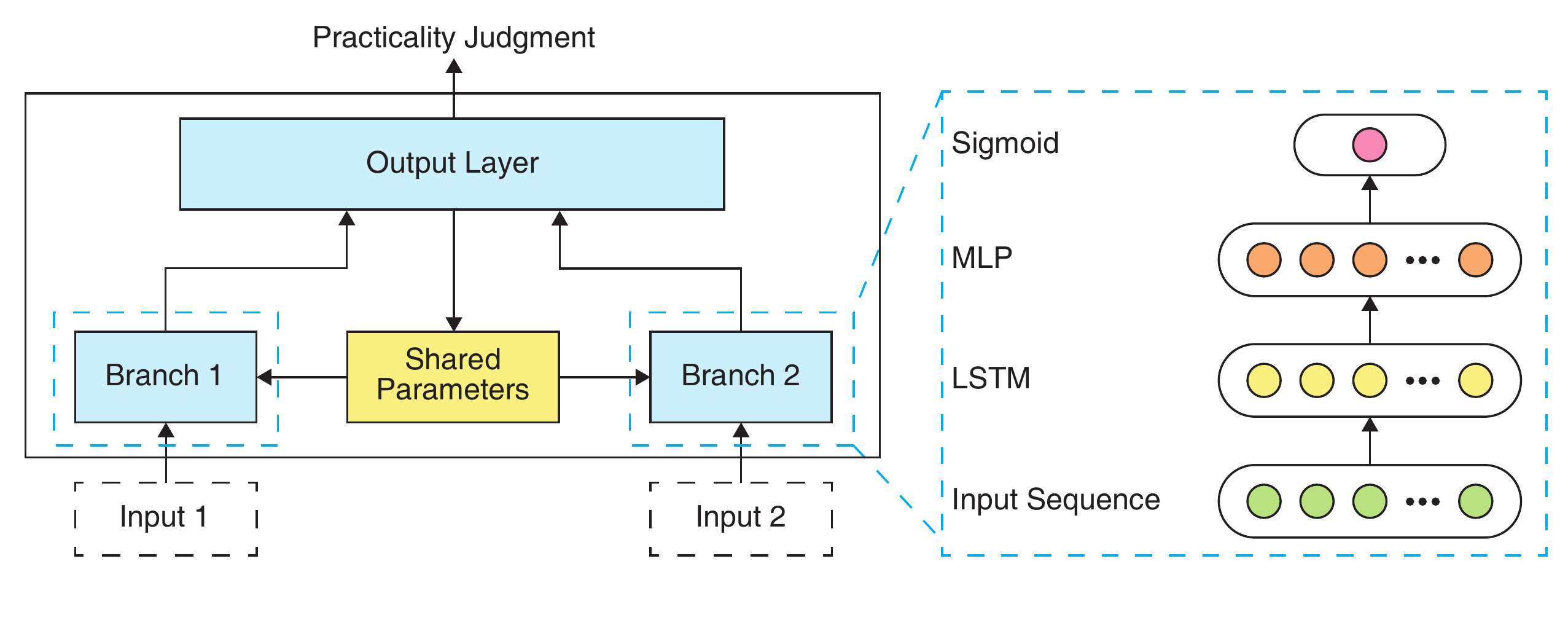}
    \caption{Siamese Network with LSTM based branch architecture.}
    \label{fig:branch}
\end{figure}

In order to learn the optimal representations of chemical reactants $\mathbf{M_1}$ and products $\mathbf{M_2}$ with RSD, we propose to use a pair-based network structure called Siamese network which has been proven as an effective framework for image matching \cite{Melekhov2016Siamese}, sequence similarity comparison tasks \cite{Mueller2016Siamese, Chopra2005Learning}. 
Since the negative reaction instances are extremely insufficient and most reported yields concentrate in a narrow range, common neural network suffers from the imbalance learning difficulty. 
The structure of Siamese network consists of two identical branches that share weights and parameters. 
Each branch poses a deep neural network for feature learning. 
In this work, we adopt Long-Short Term Memory (LSTM) Network \cite{Hochreiter1997Long} as the branch component due to its advance for sequence modeling. Figure \ref{fig:branch} shows an LSTM based branch architecture. 
The LSTM unit is defined as follows.
\begin{align}
&\mathbf{i_{t}}=\sigma (\mathbf{W}_{w}^{i}\mathbf{x}_{t}+\mathbf{W}_{h}^{i}\mathbf{h}_{t-1}+\mathbf{b}_{i}),\\
&\mathbf{f_{t}}=\sigma (\mathbf{W}_{w}^{f}\mathbf{x}_{t}+\mathbf{W}_{h}^{f}\mathbf{h}_{t-1}+\mathbf{b}_{f}),\\
&\mathbf{u_{t}}=\sigma (\mathbf{W}_{w}^{u}\mathbf{x}_{t}+\mathbf{W}_{h}^{u}\mathbf{h}_{t-1}+\mathbf{b}_{u}),\\
&\mathbf{c_{t}}=\mathbf{f}_{t}\odot \mathbf{c}_{t-1}+\mathbf{i}_{t}\odot\tanh(\mathbf{W}_{w}^{c}\mathbf{x}_{t}+\mathbf{W}_{h}^{c}\mathbf{h}_{t-1}+\mathbf{b}_{c}),\\
&\mathbf{h_{t}}=\tanh (\mathbf{c}_{t})\odot \mathbf{u}_{t},
\end{align}
where $\sigma$ stands for the sigmoid function and the $\odot$ represents element-wiselayer to form a final representation. multiplication.
$ \oplus $ denotes vector concatenation. $\mathbf{i_{t}},\mathbf{f_{t}},\mathbf{u_{t}},\mathbf{c_{t}},
\mathbf{h_{t}}$ are the input gates, forget gates, memory cells, output gates and the current states, respectively. 
Given a sequence input, the network computes the hidden sequence $\mathbf{h}_t$ by applying the formulation for each time step.

After embedding, the vectorized inputs $M_1$ and $M_2$ are separately fed to forward LSTM and backward LSTM (BiLSTM) to obtain the internal features of two directions. 
The output for each input is the concatenation of the two vectors from both
directions: $\mathbf{h}_{t} = \overrightarrow{\mathbf{h} _{t}}\parallel \overleftarrow{\mathbf{h}_{t}}$. Hence, we have the processed representations of the reactant and product with RSD, $\hat{M_1}=BiLSTM(M_1)$ and $\hat{M_2}=BiLSTM(M_2)$.
Then, our model concatenates the representation of $\hat{M_{1}}$ and $\hat{M_{2}}$ to a Multi-Layer Perception (MLP) layer to form a final representation. 
The output of the model is activated by a \emph{sigmoid} function to ensure the prediction is in [0,1].
\begin{align}
\mathbf{y} = \frac{1}{1+e^{\mathbf{x}}}
\end{align}
where $\mathbf{x} $ is the output of MLP and $\mathbf{y} $ is the prediction.

\subsection*{Training objectives}

For practicality judgment, we use binary cross entropy as the loss function.
\begin{align}
    \mathcal{L} = -\frac{1}{N}\sum_{t=1}^{n}\left [ \mathbf{y} _{t}\mathbf{log}\hat{\mathbf{y}}_{t} + (1-\mathbf{y} _{t})\mathbf{log}(1-\hat{\mathbf{y} }_{t} )\right ]
\end{align}
where $\hat{\mathbf{y}}_{t}$ denotes the prediction, $\mathbf{y}_{t}$ is the target and $t$ denotes the data index.

\section*{DATA}
The reaction data for our model evaluation has 5 sources, 
(1) a public chemical reaction dataset USPTO, 
(2) a large scale of reaction dataset extracted from reports of \emph{Chemical Journals with High Impact factors} (CJHIF), 
(3) ruled generated negative chemical reactions by Chemical.AI laboratory\footnote{\url{http://www.chemical.ai}},
(4) real failed reactions from Chemical.AI partner laboratories and 
(5) real reactions from Chemical.AI laboratories.

\subsection*{Statistics}
\begin{itemize}
\item\textbf{Positive reactions from USPTO (USPTO)}\\
This public chemical reaction dataset was extracted from the US patents grants and applications dating from 1976 to September 2016\cite{lowe_2017} by Daniel M. Lowe\cite{lowe2012extraction}. 
The portion of \emph{granted patents} contains 1,808,938 reactions described using SMILES
\footnote{\url{https://figshare.com/articles/Chemical_reactions_from_US_patents_1976-Sep2016_/5104873}}. 
Such reaction strings are composed of three groups of molecules: the reactants, the reagents, and the products, which are separated by a `$>$' sign. 
After data cleaning with RDKit\cite{rdkit}, an open-source cheminformatics and machine learning tool, it remained 269,132 items at last.

\item\textbf{Positive reactions from CJHIF(CJHIF)} \\ 
3,219,165 reactions mined from high impact factor journals\footnote{The journal list is attached in the Supplemental Material.} with \emph{reagent}, \emph{solvent} and \emph{catalyst} information, in addition with \emph{yield}. 
After data cleaning and selection, we used remaining 1,763,731 items at last.

\item\textbf{Rule-generated negative reactions from Chemical.AI (Chemical.AI-Rule)}  \\
For every product in the positive reaction sets, we adopt a set of chemical rules to generate all possible reactions which may output the respective products. Then we filter the resulted reactions by a very large known positive reaction set from Chemical.AI (which contains 20 million known reactions collected from chemical literatures and patents). Namely, all the remained unreported reactions are taken as negative reactions. Due to memory limitation, we keep 100K rule-generated negative reactions in our dataset. \\

Our idea for auto-construction of negative chemical reaction samples is actually quite intuitive, that is, we simply regard no-show reactions from any known literature are quite possibly negative ones. Only if the reference positive reaction set is large enough, we can receive quite reliable negative reasons from such filtering.

\item\textbf{Real negative reactions from Chemical.AI (Chemical.AI-Real-1)} \\ 
12,225 real failed reactions from chemical experiment record of Chemical.AI partner laboratories.
After data deduplication and canonicalization, it remained 8,797 reactions.

\item\textbf{Real reactions from Chemical.AI (Chemical.AI-Real-2)} \\
24,514 real reactions from chemical experiment record of Chemical.AI partner laboratories, in which there are 16,137 positive reactions and 8,377 negative reactions, where the productivity of negative reactions is 0\%. This data set is equally split into two parts: training set and test set.
\end{itemize}

\subsection*{Setup}
For practicality judgment, we let the two positive sets collocate with the two negative dataset to form four combinations. 
The distributions of positive and negative reaction from the train/dev/test sets are in Table \ref{tab:num}.

In our experiments, the ratio of the training set and the test set is 9:1 and 10\% of the training set is held out as development (dev) set\footnote{Dev set is used to supervise the training process in case of over-fitting or under-fitting in deep learning scenerio.}. 
For practicality judgment, since there are no negative samples from USPTO dataset, we use the positive reactions from CJHIF and USPTO to pair the real negative and rule-generated negative samples from Chemical.AI. 
For the data collected from laboratories (Chemical.AI-Real-2), we take them as the test set to examine the generalization ability of our model.
    
Considering the calculation efficiency, we specify a max length of 100 words for each SMILES sequence and apply truncating or zero-padding when needed. 
The embedding weights are randomly initialized with the uniformed distribution in the interval [-0.05, 0.05]. 

\subsection*{Evaluation metrics}
Our practicality judgment evaluation is based on the following metrics: Accuracy, Precision, Recall and F1-score. 
Four types of predictions are as shown in Table \ref{tab:out}.

Accordingly, we can calculate the performance of Accuracy, Precision, Recall and F1-score as follows.
\begin{align}
\text{Accuracy} & = \frac{\text{TP}+\text{TN}}{\text{TP}+\text{TN}+\text{FN}+\text{FP}}, \\
\text{Precision} & = \frac{\text{TP}}{\text{TP}+\text{FP}}, \\
\text{Recall} &= \frac{\text{TP}}{\text{TP}+\text{FN}}, \\
\text{F1-score} & =\frac{2 \times \text{Precision} \times \text{Recall}}{\text{Precision} + \text{Recall}}
\end{align}

\section*{EXPERIMENT}
\subsection*{Practicality Judgment}
Given input sequences, the model will output the reaction success probabilities. 
To evaluate the result, a threshold is required to distinguish from positive or negative predictions. According to our preliminary experiments, the threshold is set to 0.5\footnote{This is also the common setting for binary classification tasks  and our quantitive study shown in Figure 5 also verifies the optimal setting.}. 
The experimental result is shown in Table \ref{tab:exp}. 
We observe the positive reactions could be recognized essentially (nearly 99\%). 
Though the proportion of positive and negative cases is over 30:1, our model also ensures a high negative F1-score more than 72\%.
Besides the rule-based negative dataset have a higher Negative-F1-score. 
From the statistics of the datasets, we know the rule-based negative dataset is much bigger than the Chemical.AI-Real-1, which not only alleviates the imbalance between positive and negative examples but also increases the diversity of negative examples.

\subsection*{Generalization Ability}
In order to demonstrate the generation ability of our learning model, we report the judgment results on Chemical.AI-Real-2 dataset with different training settings in Table \ref{tab:exp2}.

As the Chemical.AI-Real-2 comes from true laboratory record, our model prediction is actually evaluated in these real chemical experiments. 
Note that these negative reactions were expected to work by experienced chemists, which means they are literally correct in chemistry rules and the chemists must encounter difficulties to predict the practicality of these reactions. Therefore, when our model gives correct practicality predication over these actually failed reactions, it means that our model performs better than human experts in these cases. 

As we know, it is difficult to get sufficient enough failed reactions because they are rarely reported in literature. Meanwhile negative examples are indispensable in discriminative machine learning on these two types of reactions. Here we show that rule-generated negative reactions for training set building may yield remarkable prediction accuracy in real chemical reaction records by considering that the rule negative samples perform best among all training settings in Table \ref{tab:exp2}. This opens a new way in the research of chemistry reaction prediction.

To have an insight of how the thresholds affect the model performance on the Chemical.AI-Real-2 dataset, we record the predication results by ranging the thresholds from 0.1 to 1 with step=0.1. The visualization results are shown Figure \ref{fig:thred}, which shows the best performance when the threshold is 0.5.

Figure \ref{fig:roc} illustrates the ROC (Receiver Operating Characteristic) curve which relates to the diagnostic ability of a binary classifier system, and the Area under the Curve (AUC) of ROC is 80.90\%, which means this model can perform quite well when the threshold sets rightly.

\subsection*{Incremental Experiment}

Different datasets may have different statistical distribution characteristics on reaction types. To fully examine the capacity of our model, we conduct a series of incremental experiments by mixing a small part of different dataset to the origin one and using the rest as test set. 

We divided the Chemical.AI-Real-2 dataset into two parts, the incremental set and test set in the ratio of 1:1. Than we add different sized parts of the incremental set with ratios [0.1, 0.2, ..., 0.9, 1.0] to the training set (\textit{USPTO + Chemical.AI-Real-1}) and conduct the experiments, respectively.

\begin{figure}
    \centering
    \includegraphics[width=1\textwidth]{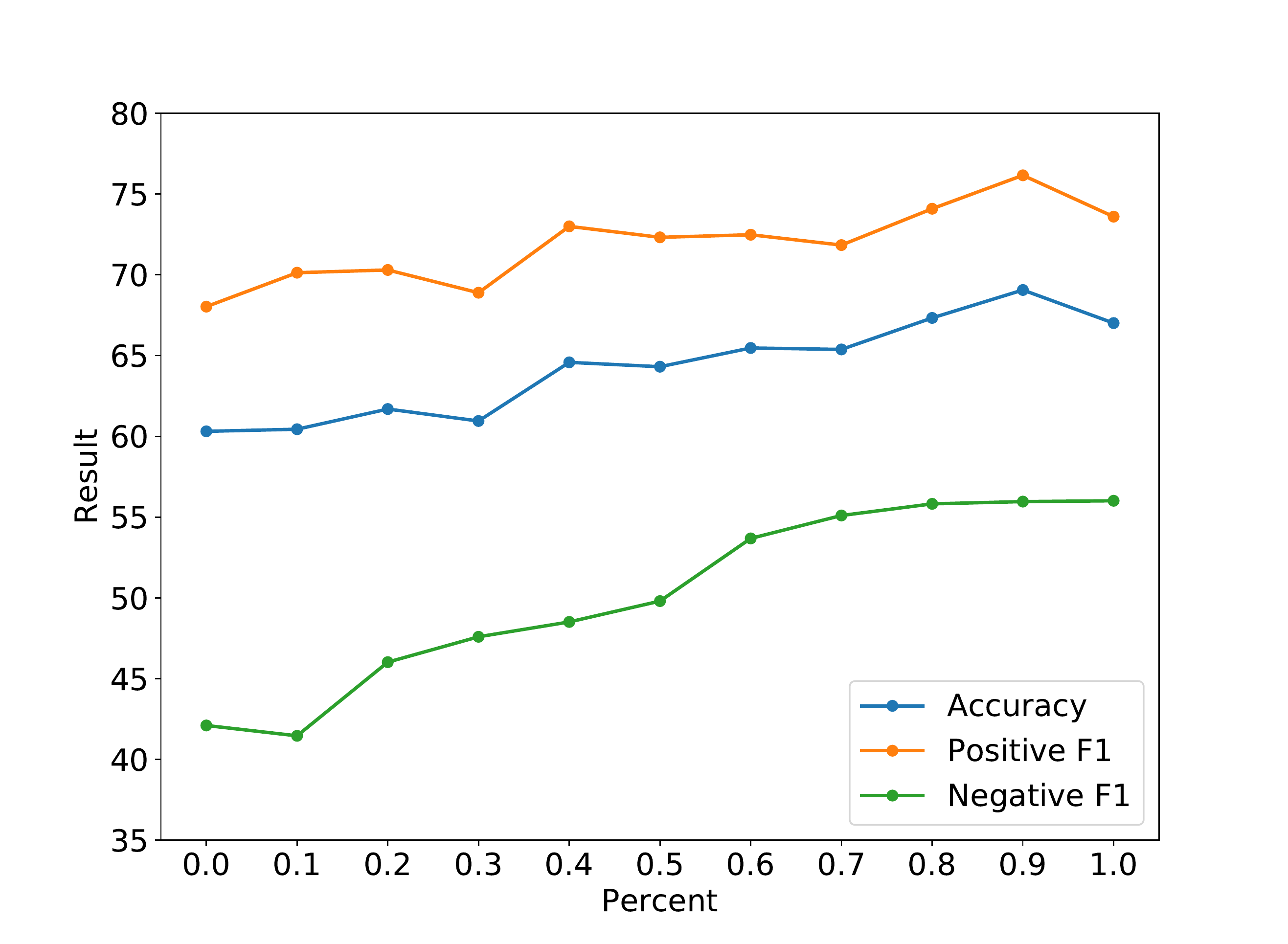}
    \caption{The result in our incremental experiment. }
    \label{fig:increase}
\end{figure}

\begin{figure}
    \centering
    \includegraphics[width=1\textwidth]{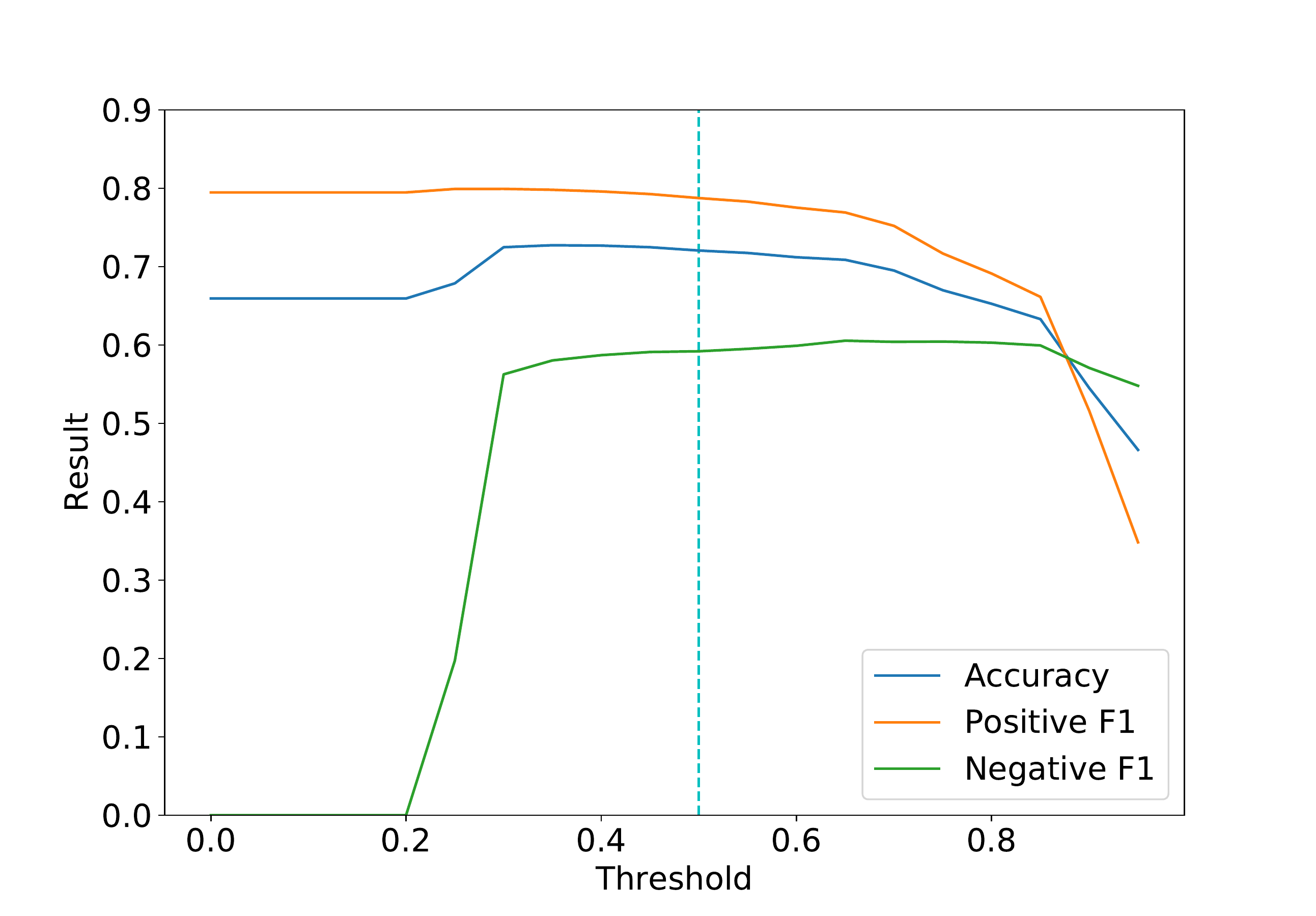}
    \caption{Threshold effect of our model on Chemical.AI-Real-2 dataset. }
    \label{fig:thred}
\end{figure}

\begin{figure}
    \centering
    \includegraphics[width=1\textwidth]{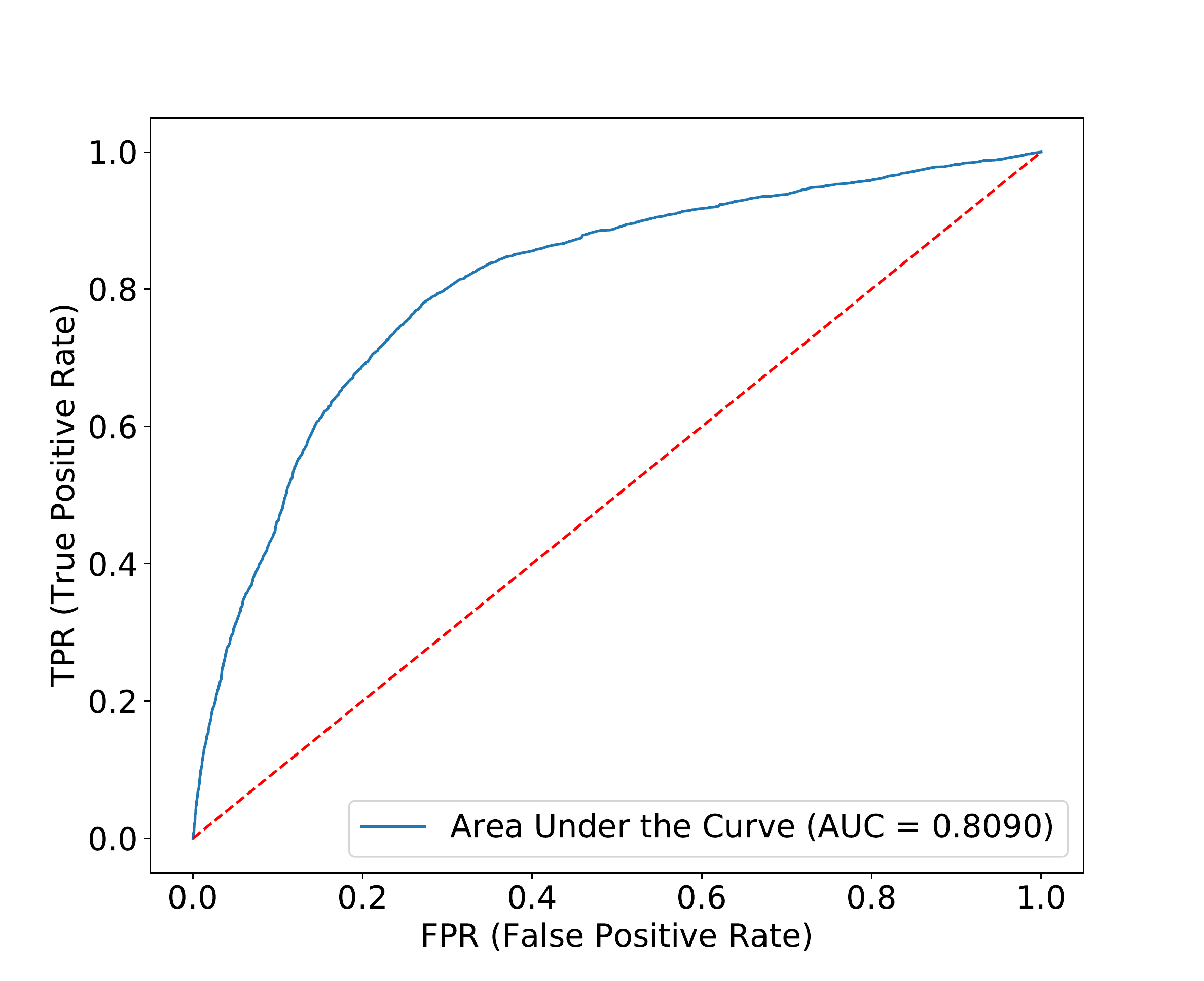}
    \caption{ROC Curve of our model on Chemical.AI-Real-2 dataset.}
    \label{fig:roc}
\end{figure}

The results in Figure \ref{fig:increase} show that even there is only a small amount of data added into the training set as the same source as the test set, the judgment results will be improved greatly.

\subsection*{DLG Segmentation}
The adopted unsupervised tokenization over the SMILES text is based on a set of \textit{“words”} with significant DLG scores in terms of the goodness measure methods. 
Despite its usefulness in our computational process, we also observe their chemical meaning. Table \ref{tab:dlg_smiles} lists a small part of the \textit{“words”} with high DLG scores. 
For example, it is not strange to any chemists that the structure of metal complex is the key to many organic reactions and in the words list, we find the number and the metallic element are always put in a same fragment which means that the ligands' position information attaching to the metallic element is useful for a better and more accurate embedding representation in our model. 
At the meanwhile, most of ordinary functional groups are also in the same fragment, like \verb|C=C|, \verb|C#C|, \verb|C=O| and \verb|C#N|, which means the model regards them as a group to process the reaction like what organic chemists do in their research. 
We also find that the ring structure in a molecule is always divided in different fragments.
Though seemingly irrational, the model actually recognizes different functional parts in a ring for more targetedly processing, which is indeed helpful to extract the reaction pattern in the later process.

\subsection*{Ablation Experiment}
We investigate the effect of different features in our model by removing them one by one. 
As shown in Table \ref{tab:analysis} all the features contribute to the performance of our final system. If we remove either \emph{RSD} or \emph{DLG Tokenizaton}, the performance drops. This result indicates that both features play an important and complementary role in the feature representation.

\subsection*{Recurrent Neural Network Types}

We also compare Siamese network with the different standard recurrent neural networks - LSTM, BiLSTM, GRU and BiGRU, and the comparison of the results is demonstrated in Table \ref{tab:baseline}. 
Obviously, Siamese network outperforms all the others, especially on the negative cases, which shows Siamese network could effectively handle the data imbalance issue.

\subsection*{Significance in Chemistry}
\begin{figure}
    \centering
    \includegraphics[width=1\textwidth]{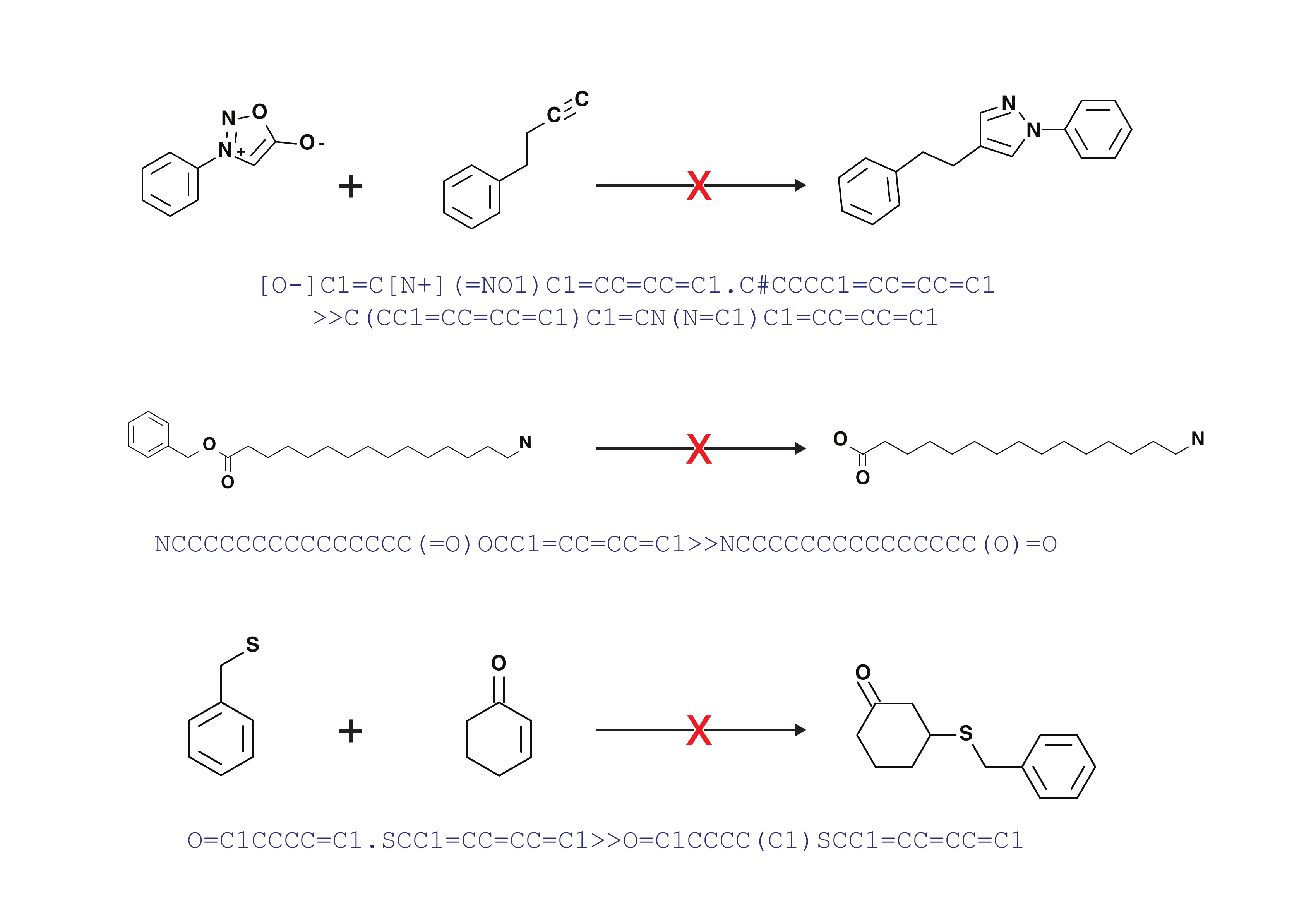}
    \caption{Positive-like cases in the test set and ``X'' means it cannot react actually.}
    \label{fig:positive}
\end{figure}

\begin{figure}
    \centering
    \includegraphics[width=1\textwidth]{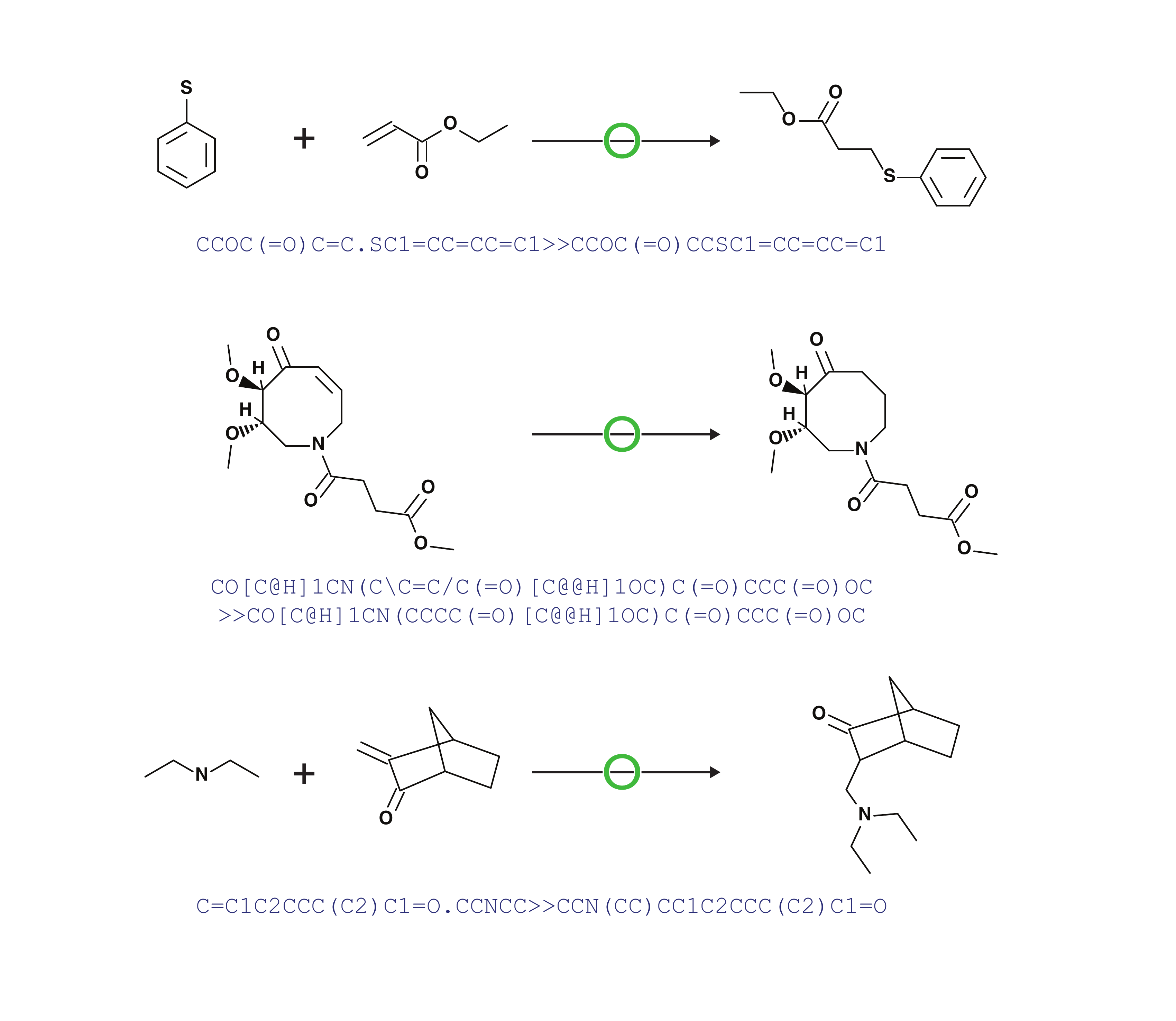}
    \caption{Negative-like cases in the test set and ``O'' means it can react actually.}
    \label{fig:negative}
\end{figure}

In practice, we have shown that our model uses the rule-based generated data as negative examples to train the model, while in the test, our model obtains significant judgment accuracies in real reaction record. It means that our model has been capable of extracting the feature of both positive and negative examples and filtering the bias introduced by rule, which shows remarkable modeling ability and will helps to improve the development of chemical engineering. 
During the test, we find that model can recognize some reactions which seems to be impractical but can react actually and some reactions which seems to be reactive but cannot react actually. Figure \ref{fig:positive} and Figure \ref{fig:negative} show such highly confused examples.


\section*{CONCLUSION}
We present a deep learning model to model real-world chemical reactions and unearth the factors governing reaction outcomes only from symbol representation of chemical information. 
In contrast to conventional methods which require massive manual features or are only evaluated on small datasets for specific reaction types, our approach is much more simple, end-to-end and effective. 
Especially, we use a rule-based method to generate unlimited negative samples, and the results evaluated on real reaction records show satisfactory judgment performance. 
In a distinctive perspective, this work reveals the great potential to employ deep learning method to help chemists judge the practicality of chemical reactions and develop more efficient experimental strategies to reduce the cost of invalid experiments. 
The resultant model can be used more than practicality judgment, but has a potential to help effective synthesis route design which has been an ongoing task in our current study and chemical practice.\footnote{Our prediction models have been online, the link is \url{http://bcmi.sjtu.edu.cn/~dl4chem}}

\bibliography{scibib}

\bibliographystyle{Science}




\clearpage

\clearpage

\begin{table}
    \centering
    
    {
    \begin{tabular}{l|l||l|l}
        \hline \hline
        \textbf{SMILES} & \textbf{Name} & \textbf{SMILES} & \textbf{Name} \\
        \hline
        \verb|CC| & ethane & \verb|[OH3+]| & hydronium ion \\
        \verb|O=C=O| & carbon dioxide & \verb|[2H]O[2H]| & deuterium oxide \\
        \verb|C#N| & hydrogen cyanide & \verb|[235U]| & uranium-235 \\
        \verb|CCN(CC)CC| & triethylamine & \verb|F/C=C/F| & E-difluoroethene \\
        \verb|CC(=O)O| & acetic acid & \verb|F/C=C\F| & Z-difluoroethene \\
        \verb|C1CCCCC1| & cyclohexane & \verb|N[C@@H](C)C(=O)O| & L-alanine \\
        \verb|c1ccccc1| & benzene & \verb|N[C@H](C)C(=O)O| & D-alanine \\

        \hline\hline
    \end{tabular}
    }
    \caption{\label{tab:smiles} Examples of SMILES}
\end{table}

\begin{table}
    \centering
    
    {\small
        \begin{tabular}{p{4.5cm}|p{10.5cm}}
            \hline
            \hline
            \textbf{Step} & \textbf{Example (reactants $>$ reagents $>$ products)}\\
            \hline
            $1)$ Original string & \texttt{[C:1]([C:3]1[CH:8]=[CH:7][CH:6]=[CH:5][C:4]=1 [OH:9])\#[N:2].[CH2:10]([CH:12]1[O:14][CH2:13] 1)Cl>N1CCCCC1>[O:14]1[CH2:13][CH:12]1[CH2:10] [O:9][C:4]1[CH:5]=[CH:6][CH:7]=[CH:8][C:3]=1 [C:1]\#[N:2]} \\
            \hline
            $2)$ Atom-mapping removal and canonicalization & \texttt{ClCC1CO1.N\#Cc1ccccc1O>N1CCCCC1>N\#Cc1ccccc1OC
            C1CO1}\\
            \hline
            $3)$ Tokenization atom-wise & \texttt{Cl C C 1 C O 1 . N \# C c 1 c c c c c 1 O > N 1 C C C C C 1 > N \# C c 1 c c cc c 1 O C C 1 C O 1}\\
            \hline
            $4)$ DLG segmentation & \texttt{ClC C 1 CO1. N\#Cc 1cccc c1 O > N1CC CCC 1> N\#Cc 1cccc c1 O C C 1 C O 1 }\\
            \hline
            \hline
        \end{tabular}
    }
    \caption{\label{tab:step}Data processing steps. The tokens are separated by a space and individual molecules by a point token.}
\end{table}

\begin{table}
    \centering
    {\small
        \begin{tabular}{l|l|r|r|r}
            \hline
            \hline
            \textbf{Data} & \textbf{Case} & \textbf{train} & \textbf{dev} & \textbf{test} \\
            \hline
            \multirow{2}{*}{CJHIF + Chemical.AI-Real-1} & Positive  & 1,406,259 & 156,251   & 173,624 \\
                                                        & Negative  & 7,178     & 798       & 874 \\
            \hline
            \multirow{2}{*}{USPTO + Chemical.AI-Real-1} & Positive  & 217,992   & 24,221    & 26,919 \\
                                                        & Negative  & 7,176     & 797       & 877 \\
            \hline
            \multirow{2}{*}{CJHIF + Chemical.AI-Rule}   & Positive  & 1,428,673 & 158,741   & 89,948 \\
                                                        & Negative  & 158,689   & 17,632    & 10,052 \\
            \hline
            \multirow{2}{*}{USPTO + Chemical.AI-Rule}   & Positive  & 217,799   & 24,200    & 90,221    \\
                                                        & Negative  & 24,421    & 2,713     & 9,779     \\
            \hline
            \multirow{2}{*}{Chemical.AI-Real-2}          & Positive  & 8,069     & -         & 8,082    \\
                                                        & Negative  & 4,202     & -         & 4,175     \\
            \hline
            \hline
        \end{tabular}
    }
    \caption{\label{tab:num} Distributions of positive and negative reaction from the train/dev/test sets in four combinations.}
\end{table}

\begin{table}
    \centering
        \begin{tabular}{c|c|c}
            \hline\hline
             & Predicted Positive & Predicted Negative \\
            \hline
            True Positive & TP & FN \\
            True Negative & FP & TN\\
            \hline\hline
        \end{tabular}
        \caption{\label{tab:out} Possible prediction results}
\end{table}

\begin{table}
    \small
    \centering
    {
        \begin{tabular}{l|c||l|c}
            \hline \hline
            \textbf{Word} & \textbf{DLG Score} & \textbf{Word} & \textbf{DLG Score} \\
            \hline
            \verb|[Rh]789%10| & 78.18 & \verb|[Ru++]5678| & 74.61 \\
            \verb|Mo+6]89%10| & 68.77 & \verb|3[Zn++]579| & 66.74 \\
            \verb|Mg]Br)cc1.| & 52.68 & \verb|ccc3)[Ru++| & 49.01 \\
            \verb|C#C[Mg]Br| & 47.60 & \verb|C=OBr[Mg]| & 40.02 \\
            \verb|\C=C/I| & 20.09 & \verb|(C#N)| & 7.68 \\
            \hline\hline
        \end{tabular}
    }
    \caption{\label{tab:dlg_smiles} Examples of SMILES words from DLG segmentation}
\end{table}

\begin{table}[h]
    \centering
    {\footnotesize
        \begin{tabular}{p{100pt}|c|c|c|c|c||c|c|c|c}
            \hline
            \hline
            \multicolumn{2}{c}{} & \multicolumn{4}{|c||}{\textbf{USPTO + Chemical.AI-Real-1}} & \multicolumn{4}{|c}{\textbf{USPTO + Chemical.AI-Rule}} \\
            \hline
            Features & Case & Precision & Recall & F1 & Acc & Precision & Recall & F1 & Acc \\
            \hline
            \multirow{2}{*}{Full}               & P & 98.83 & 99.21 & \textbf{99.02} & \multirow{2}{*}{97.92}   & 97.89 & 96.72 & \textbf{97.30} & \multirow{2}{*}{96.26}\\
                                                & N & 72.45 & 63.83 & \textbf{67.88} &                          & 91.47 & 93.94 & \textbf{92.68} & \\
            \hline
            \multirow{2}{*}{w/o \emph{RSD}}  & P & 98.81 & 98.80 & 98.81 & \multirow{2}{*}{97.78}    & 97.21 & 96.54 & 96.87 & \multirow{2}{*}{95.62}\\
                                                        & N & 63.27 & 63.63 & 63.45 &                           & 90.58 & 92.31 & 91.44 & \\
            \hline
            \multirow{2}{*}{w/o \emph{Tokenization}}    & P & 98.58 & 99.01 & 98.79 & \multirow{2}{*}{97.53}    & 96.69 & 92.33 & 94.46 & \multirow{2}{*}{93.34}\\
                                                        & N & 64.87 & 56.21 & 60.23 &                           & 84.56 & 91.53 & 86.31 & \\
            \hline
            w/o \emph{RSD} \&  & P & 98.68 & 98.61 & 98.65 & \multirow{2}{*}{97.32}    & 95.75 & 93.58 & 94.65 &  \multirow{2}{*}{93.31}\\
            \emph{Tokenization}      & N & 58.28 & 59.41 & 58.84 &                           & 83.77 & 88.87 & 86.24 & \\   
            \hline
            \hline
        \end{tabular}
    }
    \caption{\label{tab:analysis} Ablation study for practicality judgment(F1 Score on Positive case / Negative case) (\%)}
\end{table}

\begin{table}
    \centering
    {\small
        \begin{tabular}{l|l|r|r|r|r}
            \hline
            \hline
            \textbf{Model} & \textbf{Case} & \textbf{Precision} & \textbf{Recall} & \textbf{F1-score} & \textbf{Accuracy}\\
            \hline
            \multirow{2}{*}{Siamese}    & Positive  & 98.83 & 99.21 & \textbf{99.02} & \multirow{2}{*}{97.92} \\
                                        & Negative  & 72.45 & 63.83 & \textbf{67.88} \\
            \hline
            \multirow{2}{*}{LSTM}       & Positive  & 98.78 & 98.92 & 98.85 & \multirow{2}{*}{97.77} \\
                                        & Negative  & 65.24 & 60.49 & 63.83   \\
            \hline
            \multirow{2}{*}{BiLSTM}     & Positive  & 98.87 & 99.04 & 98.96 & \multirow{2}{*}{97.97} \\
                                        & Negative  & 68.87 & 65.34 & 67.06   \\
            \hline
            \multirow{2}{*}{GRU}        & Positive  & 99.13 & 98.53 & 98.83 & \multirow{2}{*}{97.74} \\
                                        & Negative  & 61.92 & 73.43 & 67.19   \\
            \hline
            \multirow{2}{*}{BiGRU}      & Positive  & 98.83 & 99.18 & 99.01 & \multirow{2}{*}{97.93} \\
                                        & Negative  & 71.78 & 64.08 & 67.71   \\
            \hline
            \hline
        \end{tabular}
    }
    \caption{\label{tab:baseline} Comparison of F1-score for practicality judgment on USPTO + Chemical.AI-Real-1(\%)}
\end{table}

\begin{table*}[h]
    \centering
    {\small
        \begin{tabular}{l|c|c|c|c|c}
            \hline\hline
            \textbf{Data} & \textbf{Case} & \textbf{Precision} & \textbf{Recall} & \textbf{F1-score} & \textbf{Accuracy} \\
            \hline
            \multirow{2}{*}{CJHIF + Chemical.AI-Real-1}  & P  & 99.82 & 99.95 & 99.88 & \multirow{2}{*}{99.76}\\
                                                        & N  & 86.09 & 63.73 & 72.24 \\
            \hline
            \multirow{2}{*}{USPTO + Chemical.AI-Real-1}  & P  & 98.83 & 99.21 & 99.02 & \multirow{2}{*}{97.92}\\
                                                        & N  & 72.45 & 63.83 & 67.88 \\
            \hline
            \multirow{2}{*}{CJHIF + Chemical.AI-Rule}  & P  & 96.19 & 99.91 & 98.02 & \multirow{2}{*}{97.96}\\
                                                        & N  & 95.15 & 75.98 & 84.49 \\
            \hline
            \multirow{2}{*}{USPTO + Chemical.AI-Rule}  & P  & 97.89 & 96.72 & 97.30 & \multirow{2}{*}{98.97}\\
                                                        & N  & 91.47 & 93.94 & 92.68 \\
            \hline\hline
        \end{tabular}}
    \caption{\label{tab:exp} Results for practicality judgment (\%)}
\end{table*}

\begin{table*}[h]
    \centering
    {\small
        \begin{tabular}{l|c|c|c|c|c}
            \hline\hline
            \textbf{Training Data} & \textbf{Case} & \textbf{Precision} & \textbf{Recall} & \textbf{F1-score} & \textbf{Accuracy} \\
            \hline
            \multirow{2}{*}{CJHIF + Chemical.AI-Real-1}  & P  & 66.00 & 85.81 & 74.61 & \multirow{2}{*}{61.49}\\
                                                        & N  & 34.42 & 14.42 & 20.32 \\
            \hline
            \multirow{2}{*}{USPTO + Chemical.AI-Real-1}  & P  & 66.19 & 26.31 & 37.65 & \multirow{2}{*}{42.98}\\
                                                        & N  & 37.65 & 34.15 & 73.99 \\
            \hline
            \multirow{2}{*}{CJHIF + Chemical.AI-Rule}  & P  & 72.03 & 79.05 & 75.38 & \multirow{2}{*}{\textbf{64.75}}\\
                                                        & N  & 50.01 & 40.57 & 44.80 \\
            \hline
            \multirow{2}{*}{USPTO + Chemical.AI-Rule}  & P  & 70.03 & 66.15 & 68.03 & \multirow{2}{*}{60.31}\\
                                                        & N  & 42.64 & 41.58 & 42.10 \\
            \hline\hline
        \end{tabular}}
    \caption{\label{tab:exp2} Results for practicality judgment in Chemical.AI-Real-2 dataset (\%)}
\end{table*}

\clearpage

\end{document}